# Deep-learning-based classification and retrieval of components of a process plant from segmented point clouds

Hyungki Kim[1,a] and Duhwan Mun[2,b]*

[1] Division of Computer Science and Engineering, Jeonbuk National University,
567, Baekje-daero, Deokjin-gu, Jeonju, Jeollabuk-do, 54896, South Korea

[2] Department of Precision Mechanical Engineering, Kyungpook National University,
2559, Gyeongsang-daero, Sangju, Gyeongsangbuk-do, 37224, South Korea

[a]hk.kim@jbnu.ac.kr, [b]dhmun@knu.ac.kr

* Corresponding Author, Tel: +82-54-530-1271, Fax: +82-54-530-1278

**Abstract**

Technology to recognize the type of component represented by a point cloud is required in the reconstruction process of an as-built model of a process plant based on laser scanning. The reconstruction process of a process plant through laser scanning is divided into point cloud registration, point cloud segmentation, and component type recognition and placement. Loss of shape data or imbalance of point cloud density problems generally occur in the point cloud data collected from large-scale facilities. In this study, we experimented with the possibility of applying object recognition technology based on 3D deep learning networks, which have been showing high performance recently, and analyzed the results. For training data, we used a segmented point cloud repository about components that we constructed by scanning a process plant. For networks, we selected the multi-view convolutional neural network (MVCNN), which is a view-based method, and PointNet, which is designed to allow the direct input of point cloud data. In the case of the MVCNN, we also performed an experiment on the generation method for two types of multi-view images that can complement the shape occlusion of the segmented point cloud. In this experiment, the MVCNN showed the highest retrieval accuracy of approximately 87%, whereas PointNet showed the highest retrieval mean average precision of approximately 84%. Furthermore, both networks showed high recognition performance for the segmented point cloud of plant components when there was sufficient training data.

# 1. Introduction

Laser scanning is a general method for acquiring as-is 3D shape data from large-scale facilities, cultural heritage buildings, or indoor environments [1]. The point clouds obtained from large-scale facilities such as buildings or process plants are as follows. First, even if the point cloud was obtained by performing laser scanning at various locations, it is difficult to obtain a point cloud that expresses the 3D shapes of all areas of the facility due to a limitation in the line of sight (LOS). Second, the density of points in the point cloud decreases as the distance from the laser scanner increases.

The process of generating a 3D model from scan data consists of point cloud collection, preprocessing, and 3D modeling [2]. In the point cloud collection process, point clouds are obtained by performing laser scanning at various locations of the site. The obtained point clouds are registered as large point cloud on a single coordinate system after undergoing a preprocessing process [3]. In the 3D modeling stage, 3D models having high-level meaning information such as the building information model (BIM) [4] or 3D computer-aided design (CAD) model [5] are created based on the registered point clouds.

The detailed process of 3D modeling to create a BIM or a plant 3D CAD model from the registered point clouds is as follows. First, the points comprising the input point cloud are grouped by similarity and divided into several segmented point clouds [6]. Then, the components represented by each separate segmented point cloud are identified. For example, it is verified if a specific segmented point cloud represents a door, window, or wall in the case of buildings or if it represents a pipe, valve, or flange in the case of a plant. 3D CAD systems in the building or plant field provide a catalog of 3D shapes for various components. Therefore, when a component represented by the segmented point cloud is identified, the corresponding catalog is selected and the 3D shapes included in the catalog are placed in the same position and orientation as those of the segmented point cloud.

The core of a 3D modeling process is to identify components represented by the segmented point clouds. A segmented point cloud is composed of a list of 3D coordinates for points. In view of this, a general method of identifying a component is to find the component that has a higher similarity value by quantifying the similarity between the shape of a segmented point cloud and the shape of a component or the user can manually decide a component corresponding to a segmented point cloud. Therefore, for automation of the process of creating a 3D model from scan data, the shape-similarity-based retrieval method that finds the component that has the most similar shape when a segmented point cloud is input is applied [7, 8].

The shape-similarity-based retrieval measures dissimilarity by extracting major features from the 3D shapes of compared objects, storing them in shape descriptors, and then calculating the distance between the shape descriptors. In general, a shorter distance indicates a low dissimilarity. Shape descriptors are classified into global shape descriptors [9,10] and local shape descriptors [11]. The point clouds obtained from a large-scale facility are likely to have only the point data for partial shapes and not the point data for the entire shape of the component. Therefore, the local shape descriptors are applied rather than global shape descriptors when applying the shape-similarity-based retrieval method to segmented point clouds acquired from large facilities.

When the traditional shape-similarity-based retrieval method is applied to the segmented point cloud obtained from large-scale facilities, the retrieval accuracy is around 60–70%, which is not high, due to the incompleteness and ambiguity of segmented point clouds and the noise included in the segmented point clouds. Consequently, when a 3D model is created from point clouds, it is difficult to automate the process of identifying components corresponding to segmented point clouds without user intervention.

To solve this problem, the present study proposes a method of identifying plant components from segmented point clouds by using a 3D deep learning technology. To that end, we first explain the segmented point cloud repository constructed for the fitting components of the process plant. Next, we propose a method of deciding the viewpoint that must be input for rendering images for segmented point clouds when applying a view-based deep learning model.

Finally, we discuss the result of identifying components represented by segmented point clouds by applying a view-based deep learning model and a point-based deep learning model.

## 2. Review of related works
### 2.1. Shape-similarity-based retrieval

The similarity comparison methods for 3D shapes are classified based on the shape descriptor expression method into the feature-based, graph-based, and geometry-based methods [8].

In the feature-based method, the geometric properties of 3D shapes are calculated and compared. Representative examples include global features such as volume [12] and moment [13], spatial maps such as spherical harmonics (SPH) [14] and 3D Zernike [15], and local features such as point signature [16] and persistent feature histogram [17].

In the graph-based method, the geometric meanings of the shape are extracted from the graph showing the relationship between elements of the 3D shape and compared. The representative graphs used in this method are the model [18], Reeb [19], and skeleton graphs [20].

The geometry-based methods are divided into the view-based [21], volumetric-error-based [22], weighted-point-set-based [23], and deformation-based methods [24]. The view-based method generates a 2D image set for each model in the database in advance and receives 2D sketches as direct input in the query stage or generates a 2D image set for the input 3D model. Then it compares the similarity between the 2D images of the query model and the model in the database.

### 2.2. 3D-deep-learning-network-based retrieval

Unlike algorithms specialized for specific tasks, deep learning is a machine learning method based on training data expression [5]. Deep learning uses layers having nonlinear processing units. Each layer converts the expression of a specific level as a more abstract expression of the level. When a cascade of layers is configured in a deep learning model, high-level abstracted information can be derived from raw input.

The success of the deep learning technology was supported by a sufficient amount of data required for training deep neural networks. Currently, there are many 2D and 3D data repositories. For example, ImageNet [32] is a dataset of over one million images of various types and each image is attached with a label (class or type). ShapeNet [33] provides three million 3D models, and 220,000 of them are classified in 3,315 categories. Princeton Shape Benchmark [34] provides a 3D model repository for the evaluation of shape-based retrieval and analysis algorithms, which include 907 categorized models. ModelNet40 [26] contains 12,311 CAD models classified into 40 categories.

Existing studies on 3D deep learning are classified into 3D voxel (voxel), 2D image, point cloud, classification of meshes [26,27], shape segmentation [28,29], and shape reconstruction [30,31]. Classification is a technique for recognizing the types of objects (e.g., chairs, furniture) corresponding to 3D models. Shape segmentation is a technique for distinguishing subparts of the input 3D shapes or for semantically segmenting 3D shapes. Shape reconstruction is a technique for generating 3D shapes through a 3D decoder after compressing 3D shapes into latent space vectors through a 3D encoder and receiving them as input.

## 3. Retrieval system concept for process plant reconstruction

The identification of components is required in the process of building an as-built model of a plant using 3D scanning technology. The overall reconstruction process consists of generating single point cloud by registering point clouds, segmenting the point clouds according to point features, identifying the component types for segmented point clouds, and then finding the corresponding catalogs and placing them in the 3D space. This study investigates component identification based on a 3D deep learning model.

Figure 1 shows the overall structure of the plant component retrieval system. This component retrieval system interacts with a commercial plant 3D CAD system. The core modules of the component retrieval system are composed of a data extraction module for separating the segmented point clouds of components from all the registered point clouds, 3D data manager module for processing and managing the separated data in a form that can be used for deep learning, and retrieval network manager for running and managing the deep learning network for retrieval. Above these modules, there is a retrieval module for performing retrieval by interacting with the API and display/UI (user interface) module for the input and output of retrievals. At the very top, there is a system manager that manages the entire system. The data used in each module includes data models, training data, and trained weights for the network.

The data model is a document of component lists structured by referring to the catalogs of the plant 3D CAD system. The training data is segmented point cloud data to be used for the training of the deep learning network for retrieval. The trained weights for the network are weights of the trained deep learning network, which are loaded to the network to identify and retrieve components from the segmented point cloud query.

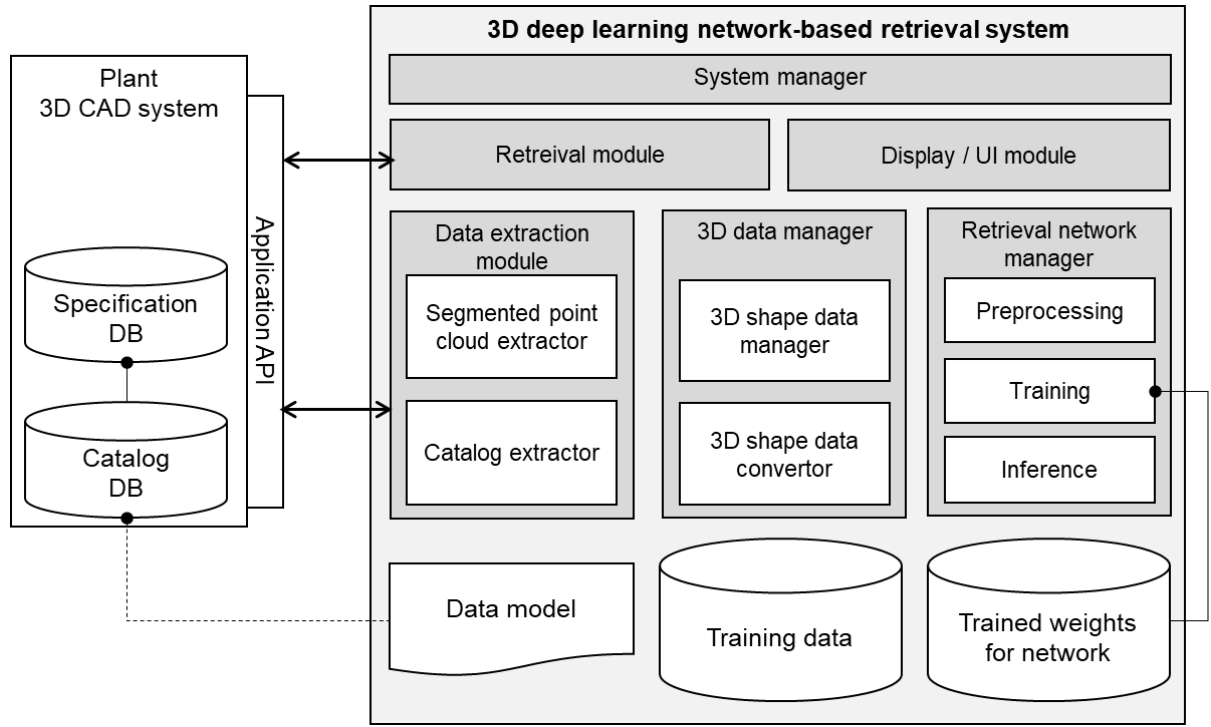

Figure 1. 3D-deep-learning-network-based retrieval system

## 4. Process plant segmented point cloud repository

To test the component retrieval performance using 3D deep learning technology, we used a segmented point cloud repository that has been constructed for the fitting components of a process plant [35]. To acquire as-built data of large-scale facilities such as a process plant, a laser scanner and other similar devices are generally used. A laser scanner can acquire the shape information of objects in the surrounding space based on the time and direction of the laser projected from the sensor and reflected by the objects. This shape information is saved as a point cloud, which is a set of 3D point coordinates.

In general, when acquiring information using a laser scanner from large facilities, point clouds are collected from multiple locations and then preprocessing is performed including noise removal. The reason that point clouds are collected from multiple locations is that there are areas where the LOS is not reached due to the straightness of light, and the areas from which highly reliable data can be acquired are limited due to the characteristics of the laser scanner. Then, all the shape information is integrated into single point cloud data through a registration process that combines the point cloud coordinates based on each sensor coordinate system into one global coordinate system.

The process plant segmented point cloud repository used in this study was created by collecting point clouds from an actual process plant and then registering and processing them. A subset of the registered point clouds corresponding to fitting components was identified by a user and then segmented using a user interface provided by the point cloud processing software. For classification criteria of the fitting components, the data model defined in [35] was used by referring to the specification catalog of a commercial plant 3D CAD system.

A total number of 4,633 segmented point clouds for 18 types of fitting components were included in the repository. Figure 2 shows the types of fitting components and the number of segmented point clouds for each type.

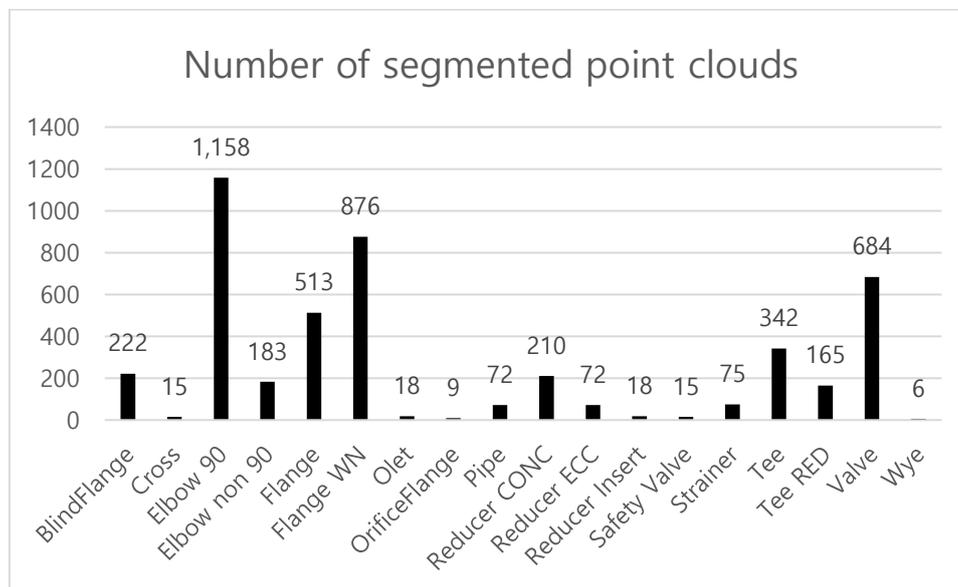

Figure 2. Segmented point clouds for components of process plant

The constructed segmented point cloud repository has the following characteristics. To achieve high repository performance, the characteristics of the point cloud data must be considered together with the following characteristics of the repository.

1) *Shape data loss due to occlusion*: A part of shape of fitting components acquired through a laser scanner have lost due to occlusion. Occlusion is caused by other components between the laser scanner and the scanned component or by self-occlusion due to the shape of the component itself. For some point clouds, the lost information from a single scanner is complemented through the registration of point clouds collected from multiple locations, but not always.

2) *Imbalance of point cloud density*: Due to the characteristics of the laser scanner, the larger the distance between the object and scanner sensor, the smaller the collectable shape information per unit area becomes. Therefore, even for the same type of components, there is a difference in the density of collected point clouds if the distance from the sensor is different. If the distance is the same, the collectable shape information per unit area is the same, but the density may vary depending on scale normalization which is required in scale-invariant retrieval.

3) *Imbalance of the number of data by component type*: There is an imbalance in the number of data by component type due to the characteristics of the process plant from which point clouds were collected, selection of the scanning location, and experience of the worker during segmentation. As shown in Figure, 2, the Wye which has the smallest quantity contains only six segmented point clouds, whereas the Elbow 90 which has the largest quantity contains 1,156 point clouds.

4) *Lack of normal and sensor position data*: The normal data associated with each point in the point cloud can be estimated by calculation through post-processing after collection or a separate process during point cloud collection. However, the point clouds of the segmented point cloud repository do not include normal data. Furthermore, the relative position data between point clouds and sensors is also lost in the process of dividing into individual point clouds through segmentation after transformation of all point clouds into one global coordinate system through registration.

## 5. Deep-learning-network-based retrieval methods

To perform retrieval tests for the point cloud repository, two types of deep-learning-network-based retrieval technologies were selected. One is a view-based method, which recognizes the shapes of 3D models by projecting them into 2D images, and the other is to configure a network to enable training by using the point clouds as input data. Among the view-based methods, we selected the multi-view convolutional neural network (MVCNN) [36], which is a widely used 3D model retrieval technology. For the point cloud input method, we selected PointNet [37].

## 5.1. MVCNN

The MVCNN converted a 3D object recognition problem into a 2D image recognition problem and showed better performance than existing methods. The existing deep-learning-based 3D recognition methods generally used an occupancy grid defined based on voxels. The voxel-based method defines a space of grids as a way to express 3D data consistently and uses the data encoding occupancy information for the 3D model shape included in each grid as input. It encodes a larger amount of data from the original shape when the 3D grid resolutions are increased, but it has a limitation in increasing the resolution due to the limited memory and computation power.

Figure 3 shows the method proposed using the MVCNN. Multiple 2D images are created by rendering a 3D object from various viewpoints, and the features are calculated through the first convolution layer ($CNN_1$) from these images. Then, the labels of objects are recognized in the second convolution layer through view pooling, which selects characteristic features that can be used for object recognition from the features of each image.

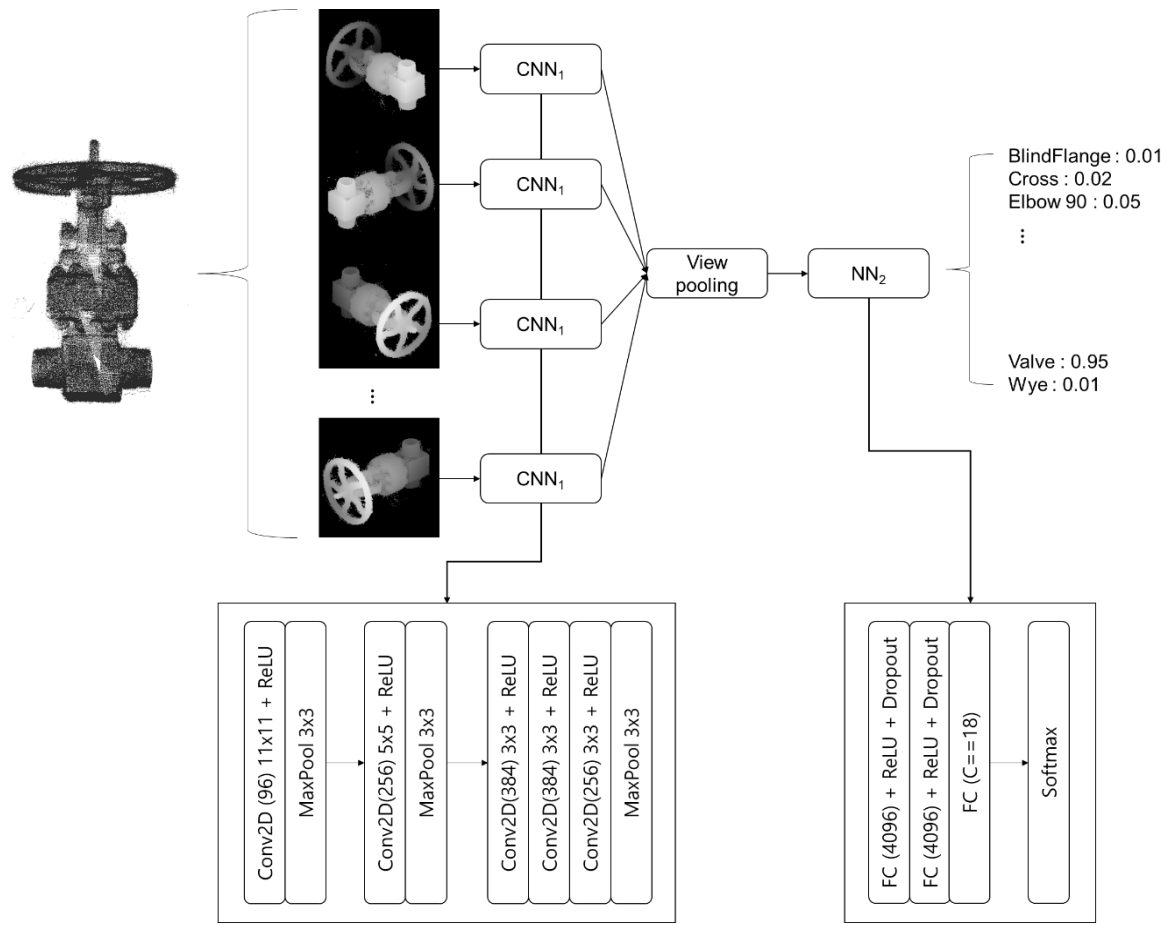

Figure 3. MVCNN network architecture

According to the authors of the MVCNN, it showed a maximum classification accuracy of 90.1% and a retrieval mean average precision performance of 80.2% in the experiment for the Princeton ModelNet dataset. This is much higher performance than the feature-based machine learning methods such as SPH (Spherical Harmonics) [38] and Light Field Descriptor [39]. Furthermore, it showed higher performance than the recognition using a simple convolutional neural network (CNN) structure with no pooling of multi-view data. The authors performed a sketch-based 3D shape retrieval experiment by applying the proposed methodology, and it also showed higher performance than the existing method.

To apply the MVCNN to segmented point cloud data, a rendering process is required to transform the segmented point clouds into multi-view images. Because the MVCNN assumed the input 3D data as 3D surface data expressed as a polygon mesh, watertight images are derived when rendering is performed using the Phong reflection model. However, in the case of point cloud data, sparse images are derived because colors are calculated only for pixels at which a virtual camera and point data intersect with each other on the image plane. Figure 4 shows an example of rendering result of the selected Blind Flange point clouds in the segmented point cloud repository. As shown in this figure, the images are formed in black, which is the background color, in the surrounding area except the pixels to which each point is projected. The sparsity appears differently depending on the density of the point cloud data and the location of the virtual camera.

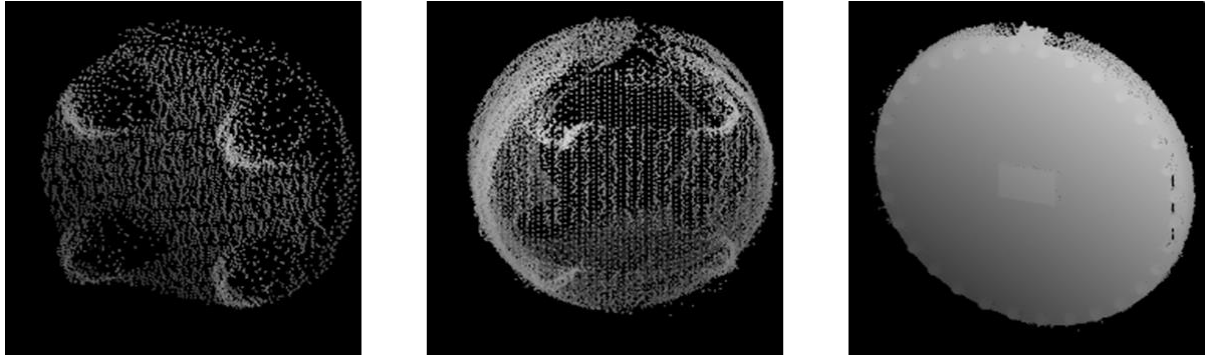

Figure 4. Rendering of point cloud for Blind Flange

To derive watertight images from point cloud data as input, a preprocessing to reconstructing point clouds into a polygon mesh is required. However, this preprocessing has disadvantages: 1) it requires additional computation time, and 2) robust reconstruction is difficult when the densities of input point clouds are diverse such as we can observe in the segmented point cloud repository and there is information loss due to occlusion. Therefore, retrieval with no reconstruction is preferred. In contrast, when reconstruction is not performed, the convolution layer used to train the image features in the CNN may not learn the correct features. Convolution operation uses the weighted sum of the central and surrounding pixel values, and it is difficult to derive meaningful features for the pixels of which all the surrounding pixel values has background color.

### 5.2. MVCNN with view selection

When using the MVCNN network in this study, in addition to the method of using multi-view images rendered in all directions as input, we also performed experiments using images generated by view selection methods based on random sample consensus (RANSAC) and based on acquisition rate [40]. The view selection method was proposed to resolve the above-mentioned problem of sparse images derived when rendering point clouds. The acquisition-rate-based view selection method showed improved retrieval performance when applied to the machine-learning-based retrieval method.

Assuming that point clouds were acquired from sensors installed at a location, in order to render the data included in a point cloud into images with no loss, the relative positions of the virtual camera and point cloud must be identical to the relative position of the sensor and point cloud. As shown in Figure 5, the laser scanner generally scans a target object by isometric projection of the laser. When rendering point clouds, if the relative position of the virtual camera and object is identical to the relative position of the laser scanner and object, all points are projected at even intervals to the image plane. If the relative positions are different, they are not projected evenly, and the information of some points is ignored by other points.

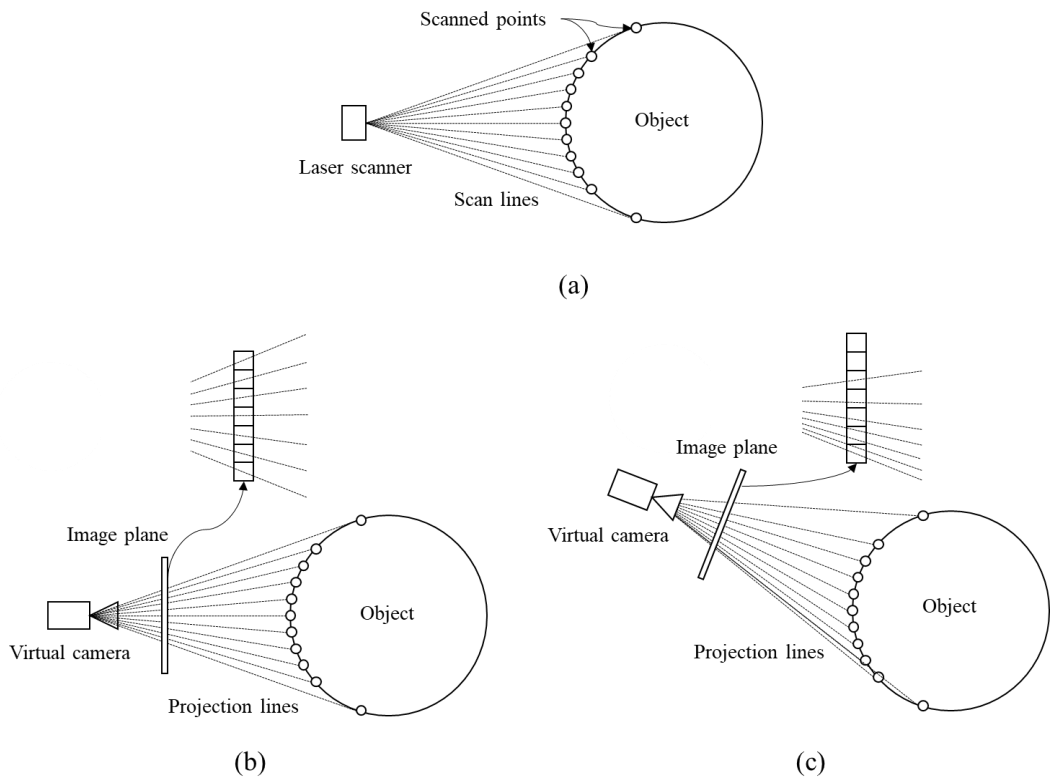

Figure 5. Effects of camera's position on rendering of point cloud (a) Position of laser scanner and target object (b) When virtual camera is located at same position as scanner (c) When virtual camera is located at different position from scanner

To examine the effect of rendering method on the deep-learning-based retrieval performance, the positions of the virtual cameras were selected using three methods. The first is the method proposed in the existing MVCNN. First, the point cloud is normalized into a unit sphere whose center is at the origin and radius is 1. Rendering is performed by placing the virtual camera at a 30º inclination from the unit sphere while increasing the azimuth angle from 0 to 360º in 30º intervals so that the origin is looked from 12 camera positions in total.

The second method is to normalize the segmented point cloud and then select the position of the virtual camera by estimating the fitted plane using RANSAC. As shown in Figure 6 (a), the fitted plane with the smallest sum of squared distances is found within the threshold from the input point cloud and then the intersection point between the normal vector of the fitted plane and the unit sphere is selected as the camera position. After that, rendering is performed with a total of 13 camera positions by increasing or decreasing the inclination and azimuth angle in equal intervals and setting the cameras to look the origin from the selected positions.

The third method is based on the acquisition rate. After normalization of the segmented point cloud, as shown in Figure 6 (b), a dodecahedron surrounding the normalized point cloud is defined. Then, the first rendering is performed using virtual cameras that look the origin from the position of the 20 vertices of the dodecahedron. Using the 20 images generated by the rendering, the acquisition rate suggested in [40] is calculated and the position with the highest acquisition rate is selected as the position of the virtual camera. The acquisition rate is a ratio of the number of pixels rendered into an image over the number of point data of the point cloud. Then, as with RANSAC, a total of 13 camera positions are selected by increasing or decreasing the inclination and azimuth angle.

Figure 7 illustrates the proposed MVCNN method and the rendering methods using view selection based on RANSAC and acquisition rate. In this figure, each triangular pyramid indicates the position and direction of the virtual camera. The proposed MVCNN method generates images evenly in all directions of the target shape, but the methods based on RANSAC and acquisition rate generate images by changing the inclination and azimuth angle around a specific position.

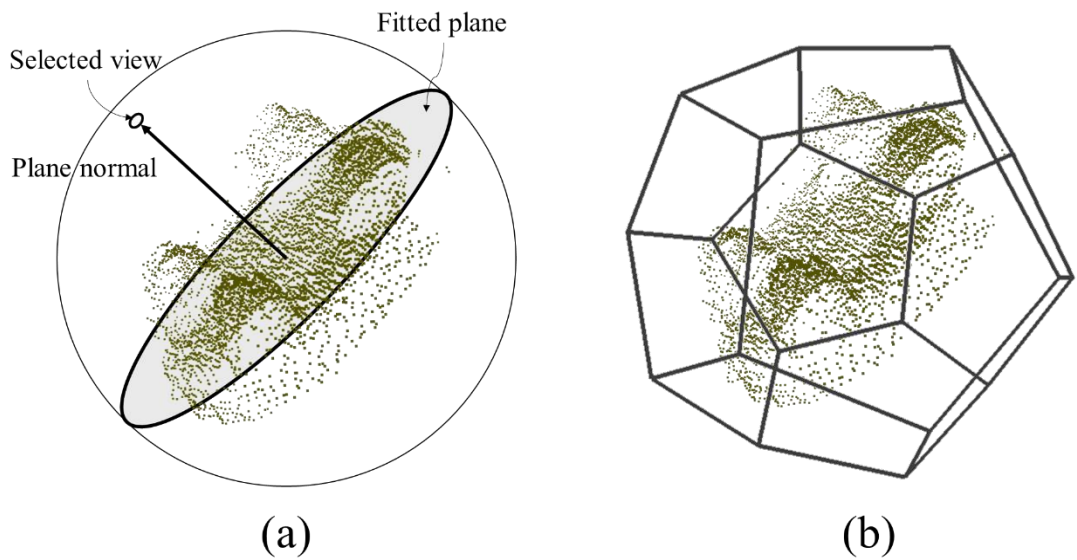

Figure 6. View selection methods: (a) RANSAC-based view selection (b) Acquisition-rate-based view selection using dodecahedron

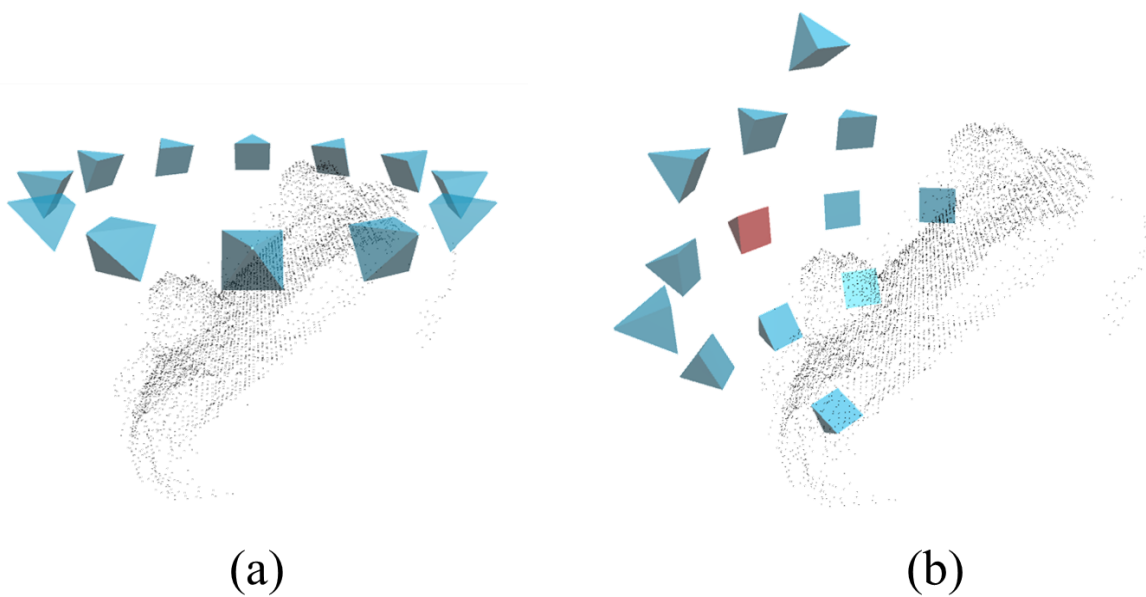

Figure 7. Positions of camera used to generate multi-view images (a) proposed method using MVCNN (b) RANSAC- and acquisition-rate-based methods

## 5.3. PointNet

The PointNet is a network structure for classification and segmentation of point cloud data. To solve the disadvantage of the above-mentioned voxel-based method, it must be possible to use the point cloud data as direct input. One characteristic of point cloud data is that it is unordered data, for which the encoding order of data cannot be specified. Even if the data were collected from the same shape, the input data can differ by the recording order of point clouds. Hence, a permutation-invariant network is required to train the same features from data with different orders in the network training process. In addition, a transform-invariant network for robust recognition is also required because the point clouds can be transformed according to the sensor position and registration.

PointNet enabled robust feature recognition using point cloud data, which is unordered input data, by using a symmetric function. It also enabled the recognition of the same features for models having different transforms by placing T-net in the middle. Figure 8 shows the overall structure of PointNet used in this study.

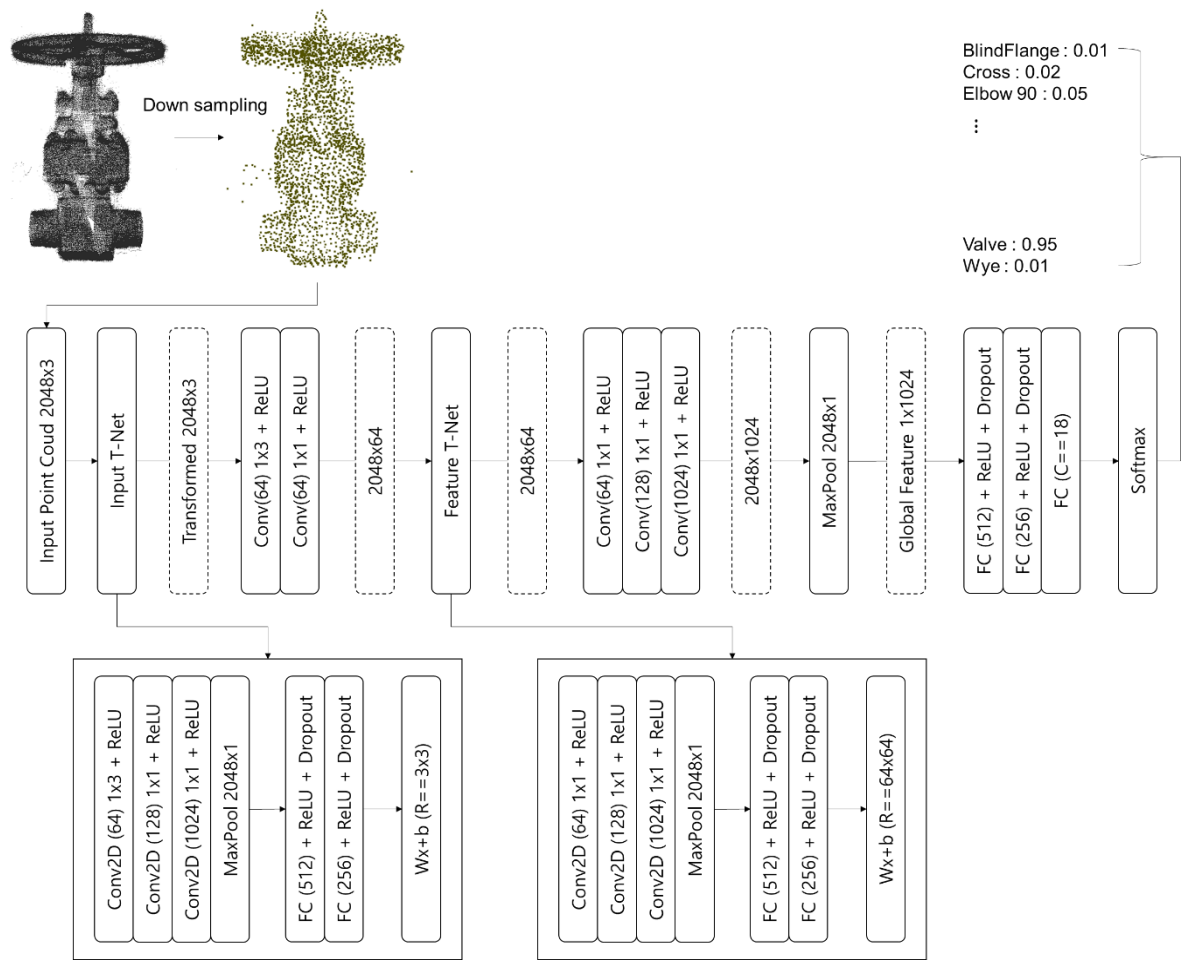

Figure 8. PointNet network architecture

## 6. Experiment
### 6.1. Input data

Figure 9 shows 18 fitting components included in the segmented point cloud repository used in the experiment on the deep-learning-based segmented point cloud retrieval. To facilitate the identification of the shape of each component, we selected segmented point clouds with low occlusion and high density. Figure 10 shows rendering examples showing the characteristics of point cloud data among the segmented point clouds of the Blind Flange type. Loss of information due to occlusion and low-density point clouds due to the distance from the sensor can be seen. In the case of the segmented point cloud containing noise, point data unrelated to fitting components was included in normalization process, resulting in excessively scaled image after rendering.

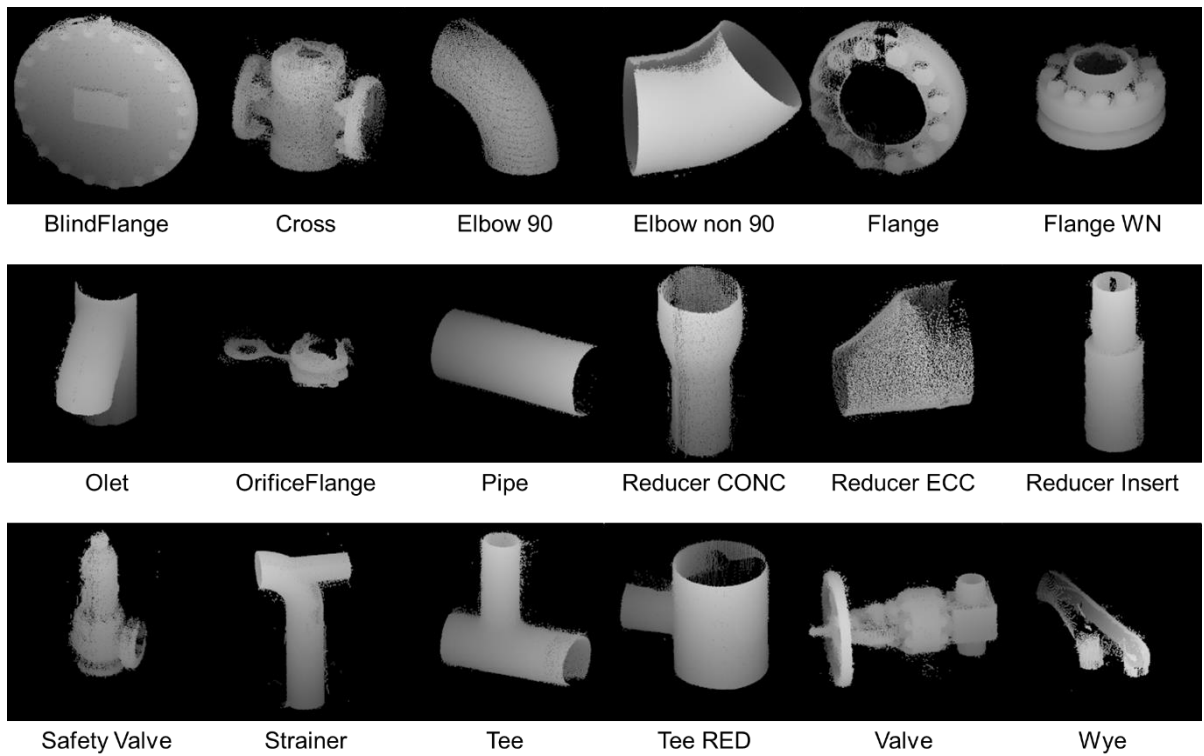

Figure 9. Eighteen types of fitting components stored in segmented point cloud repository

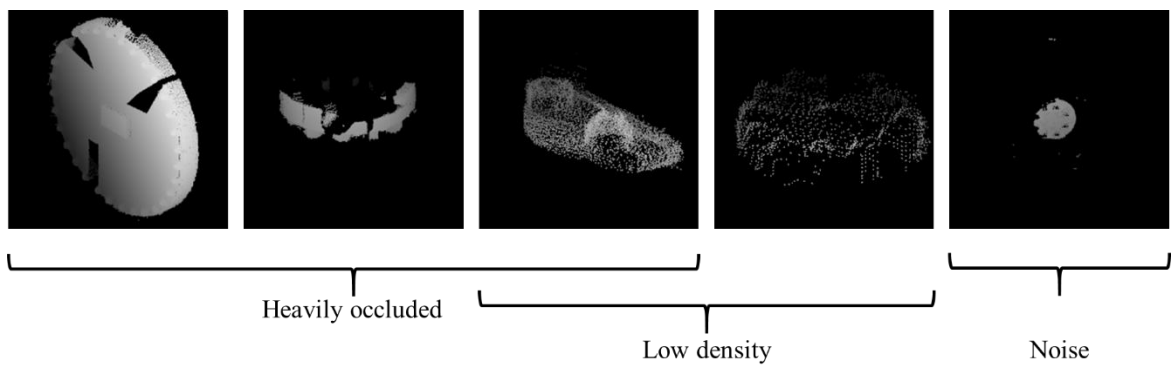

Figure 10. Occlusion, low density, and noise of segmented point clouds

Figure 11 shows the multi-view images rendered by the proposed MVCNN method, RANSAC-based method, and acquisition-rate-based method. These images were rendered with a resolution of 227 by 227 and the color of each pixel encodes the depth value of point data. A brighter color indicates a point closer to the virtual camera and a darker color indicates a point farther from the virtual camera. In the images rendered by the RANSAC-based method and the acquisition-rate-based method, the image at the center indicates the calculated location of the virtual camera, and the other images were rendered by increasing or decreasing the inclination angle and azimuth angle in 10º intervals. The proposed MVCNN method generates 12 images per point cloud, and the other two methods generate 13 images per point cloud.

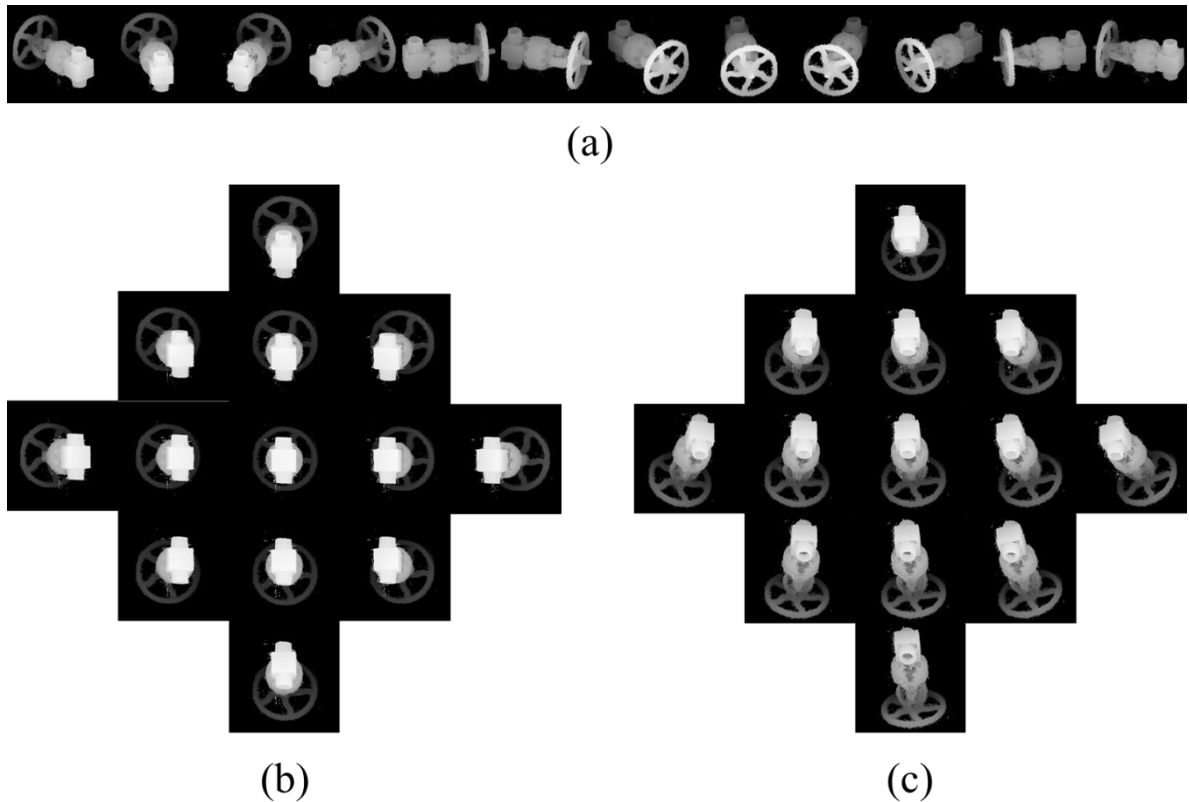

Figure 11. Rendered multi-view images used in experiments (a) Proposed MVCNN method (b) RANSAC-based method (c) acquisition-rate-based method

A total of six test cases were experimented. For five cases, the MVCNN network was used with the rendered image of the point cloud as input. For one case, the PointNet network was used with the point cloud data as input. Table 1 summarizes the test cases. For the image data, the number of individual data is determined by the total number of segmented point clouds multiplied by the number of views, but the images of the number of views are used for training and inferring single model. In Table 1, MVCNN indicates the image rendered by the method of Figure 11 (a), RANSAC indicates the image rendered by the method of (b), and acquisition rate indicates the image rendered by the method of (c). Degree 10 and Degree 40 indicate the angles between the virtual camera selected based on the RANSAC and acquisition rate and the other 12 surrounding virtual cameras.

Table 1. Six test cases of experiment

| Input data | Type | Resolution / number of points | Number of training data | Number of validation data |
|---|---|---|---|---|
| **MVCNN** | Image | 227 x 227 | 44,580 | 11,256 |
| **RANSAC (Degree 10)** | Image | 227 x 227 | 48,295 | 12,194 |
| **RANSAC (Degree 40)** | Image | 227 x 227 | 48,295 | 12,194 |
| **Acquisition rate (Degree 10)** | Image | 227 x 227 | 48,295 | 12,194 |
| **Acquisition rate (Degree 40)** | Image | 227 x 227 | 48,295 | 12,194 |
| **Sampled point cloud** | Point cloud | 2,048 | 3,715 | 938 |

## 6.2. Training

Both the MVCNN and PointNet were implemented using Tensorflow. For each type of fitting component, 80% of the point cloud data was used for training, and 20% of the data was used for validation and performance measurement. Adam optimizer was used for both the MVCNN and PointNet. In the case of PointNet, the point cloud that was generated by downsampling 2,048 points from the original point cloud was used as input. The batch sizes of 64 and 32 were used for training in the case of the MVCNN and PointNet, respectively. In addition, the learning rates of 0.001 and 0.0001 were used for the MVCNN and PointNet, respectively.

Figure 12 graphically shows the convergences of validation loss and accuracy for the validation data in the training. PointNet and the MVCNN are shown as separate graphs because they have different loss scales. It can be seen that the losses of both networks converged. The training and later performance measures were calculated on a PC with Intel i7 CPU (3.70GHz), NVIDIA GeForce GTX 1080 Ti GPU, 32GB RAM, and Windows 10 64x OS.

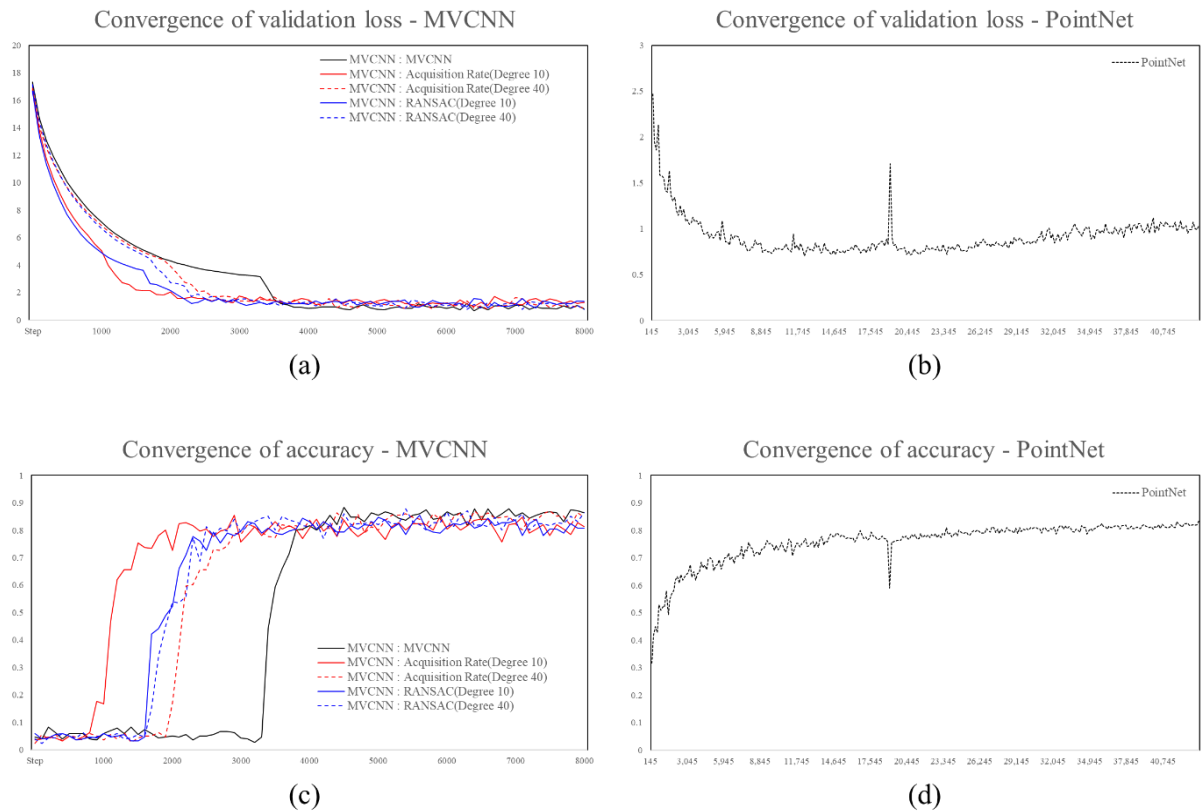

Figure 12. Validation losses and accuracy convergences of MVCNN and PointNet

## 6.3. Results

Table 2 outlines the performance results of the test cases. The highest performance value in each metric is indicated in bold font. Overall accuracy is the ratio of correct classification results for all query data. Class accuracy is the average accuracy of each class (fitting type). mAP is the average precision of each class averaged for all classes. Overall accuracy and class accuracy were the highest at 87.41% and 80.02% in the case of rendering the image by the MVCNN and training in the MVCNN network. The case of rendering the image based on the acquisition rate and training in the MVCNN network showed the second highest performance. In the case of mAP, the case of training the sampled point cloud in PointNet showed the highest performance, followed by the case of using image data based on the acquisition rate.

Table 2. Accuracy and mean average precision result from six test cases

| *Input data* | *Network* | *Overall accuracy* | *Class accuracy* | *mAP* |
|---|---|---|---|---|
| **MVCNN** | MVCNN | **87.41%** | **80.02%** | 78.19% |
| **RANSAC (Degree 10)** | MVCNN | 81.43% | 66.96% | 78.78% |
| **RANSAC (Degree 40)** | MVCNN | 81.22% | 65.53% | 80.00% |
| **Acquisition rate (Degree 10)** | MVCNN | 82.28% | 78.76% | 79.75% |
| **Acquisition rate (Degree 40)** | MVCNN | 85.06% | 71.86% | 81.19% |
| **Sampled point cloud** | PointNet | 79.98% | 73.66% | **84.35%** |

Table 3 shows the performance metrics excluding the lower 50% types, which have a small amount of data. These performance metrics were calculated excluding the retrieval results for Cross, Olet, Orifice Flange, Pipe, Reducer ECC, Reducer Insert, Safety Valve, Strainer, and Wye. For these types, fewer than 60 segmented point clouds were used for training, and fewer than 15 segmented point clouds were used for validation. The accuracy and mAP increased in general, but the tendency was the same as in Table 2.

Table 3. Accuracy and mean average precision result from six test cases when 50% types that had lower numbers of data were excluded

| *Input data* | *Network* | *Overall accuracy* | *Class accuracy* | *mAP* |
|---|---|---|---|---|
| **MVCNN** | MVCNN | **87.41%** | **82.95%** | 78.55% |
| **RANSAC (Degree 10)** | MVCNN | 82.38% | 77.07% | 79.88% |
| **RANSAC (Degree 40)** | MVCNN | 82.49% | 78.46% | 81.31% |
| **Acquisition rate (Degree 10)** | MVCNN | 83.30% | 80.20% | 81.35% |
| **Acquisition rate (Degree 40)** | MVCNN | 85.01% | 79.93% | 81.66% |
| **Sampled point cloud** | PointNet | 78.83% | 74.17% | **84.38%** |

Table 4 shows the accuracy of each class for 18 fitting types. The input data that showed the highest performance for each fitting type is indicated in bold font. The types with the largest training data (Elbow 90, Flange WN, Valve) generally showed high performance regardless of the test case. The types having less data than this showed deviations in performance depending on the input data and network. The performance was particularly low in the types of Blind Flange in the MVCNN, Flange in the sampled point cloud, Reducer CONC in acquisition rate, and Tee RED in RANSAC. These cases were marked using underline and italic font. The performance deviation was not considered for the lower 50% types that had a small amount of data.

Table 4. Overall accuracy of six test cases for each fitting component type

| | *MVCNN* | *RANSAC (Degree 10)* | *RANSAC (Degree 40)* | *Acquisition rate* | *Acquisition rate* | *Sampled point cloud* |
|---|---|---|---|---|---|---|

|  |  |  |  | (Degree 10) | (Degree 40) |  |
|---|---|---|---|---|---|---|
| **Blind Flange** | _0.80_ | 0.91 | 0.89 | **0.93** | 0.87 | 0.91 |
| **Cross** | 0.00 | **0.67** | 0.00 | 0.33 | 0.00 | 0.00 |
| **Elbow 90** | 0.96 | 0.95 | **0.97** | 0.92 | 0.91 | 0.93 |
| **Elbow non 90** | 0.68 | 0.51 | **0.76** | 0.73 | 0.65 | 0.68 |
| **Flange** | **0.77** | 0.61 | 0.72 | 0.60 | 0.75 | _0.42_ |
| **Flange WN** | **0.85** | 0.74 | 0.65 | 0.79 | **0.85** | 0.79 |
| **Olet** | 0.00 | **1.00** | **1.00** | 0.75 | 0.75 | **1.00** |
| **Orifice Flange** | **1.00** | 0.50 | 0.00 | **1.00** | **1.00** | **1.00** |
| **Pipe** | **1.00** | 0.80 | 0.80 | 0.80 | 0.93 | **1.00** |
| **Reducer CONC** | **0.93** | 0.79 | 0.81 | _0.69_ | 0.86 | 0.71 |
| **Reducer ECC** | **1.00** | 0.60 | 0.73 | 0.53 | 0.93 | 0.80 |
| **Reducer Insert** | **1.00** | 0.75 | 0.50 | 0.50 | **1.00** | 0.25 |
| **Safety Valve** | **1.00** | 0.00 | 0.67 | **1.00** | **1.00** | 1.00 |
| **Strainer** | **0.93** | 0.80 | 0.53 | 0.80 | 0.87 | **0.93** |
| **Tee** | 0.77 | **0.83** | 0.80 | 0.81 | 0.75 | 0.77 |
| **Tee RED** | 0.73 | 0.61 | _0.49_ | 0.76 | 0.58 | **0.82** |
| **Valve** | 0.99 | 0.99 | 0.99 | 0.99 | 0.98 | 0.99 |
| **Wyw** | **1.00** | 0.00 | 0.50 | 0.00 | 0.50 | **1.00** |

The reason for the lower performance of Flange in the sampled point cloud is the failure of distinguishing from the Flange WN type. When the Flange and Flange WN types were classified as the same type, the accuracy increased from 42% to 95%. Figure 13 shows the CAD models for the Flange and Flange WN types and the shapes of the segmented point clouds after actual scanning. As shown in this figure, some Flange WN models do not show a clear difference between the Flange model in its shape. PointNet showed lower recognition performance for these cases than the MVCNN.

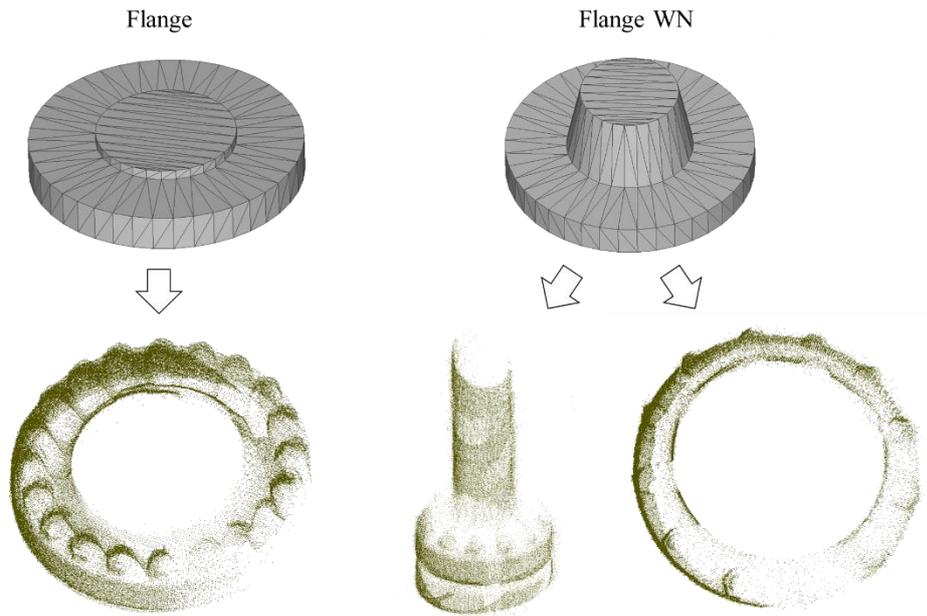

Figure 13. 3D CAD models and segmented point clouds for Flange and Flange WN type fitting components

Meanwhile, in the case of Blind Flange in the MVCNN, images are generated in all directions of the segmented point cloud. As a result, many images of the side or back of the shapes are included, as shown in Figure 14 (a). By contrast, the virtual cameras selected based on the acquisition rate usually generate front images of the shape, as shown in Figure 14 (b). This means that these images are appropriate for the classification of fitting types.

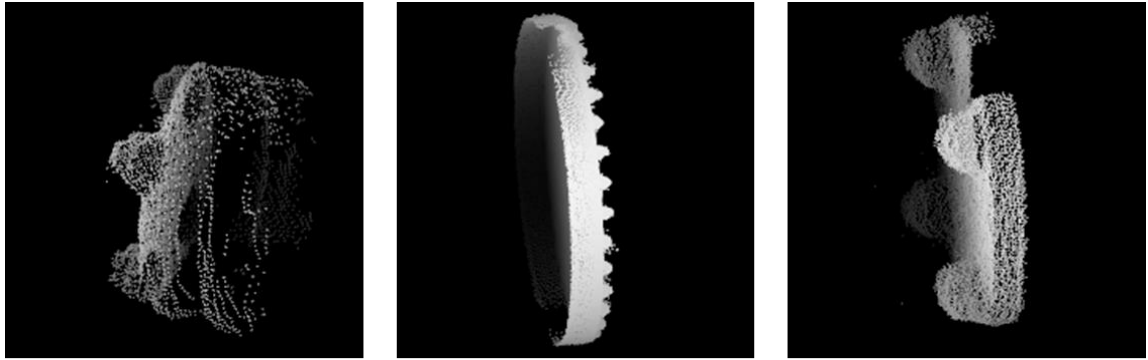

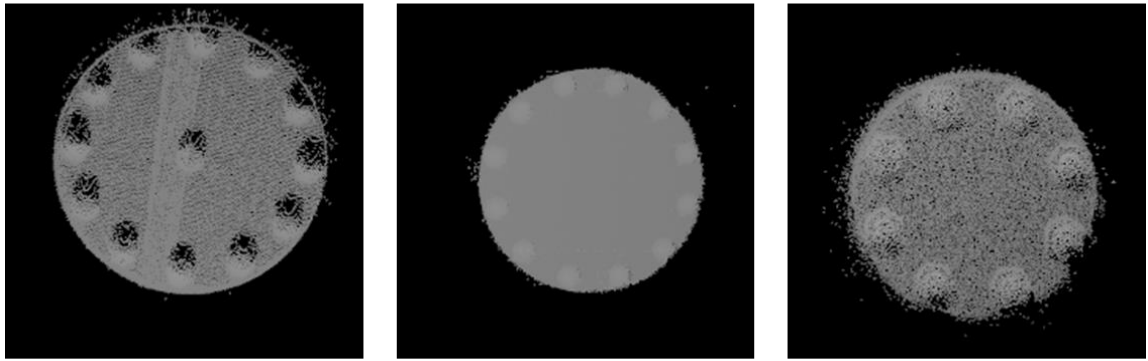

Figure 14. Rendered multi-view images for Blind Flange type fitting component (a) proposed MVCNN method (b) acquisition-rate-based method

Reducer CONC and Tee RED showed low accuracy in 10 degree intervals of acquisition rate and 40 degree intervals of RANSAC, respectively. These two fitting types commonly have relatively small datasets. Consequently, Tee RED was often misrecognized as Tee type. Reducer CONC was often misrecognized as Elbow 90 type, which has the largest amount of data and has a similar cylindrical shape.

Figure 15 shows the precision-recall curves of test cases. Figure 16 shows the precision-recall curves with the exclusion of the lower 50% models with a small amount of data.

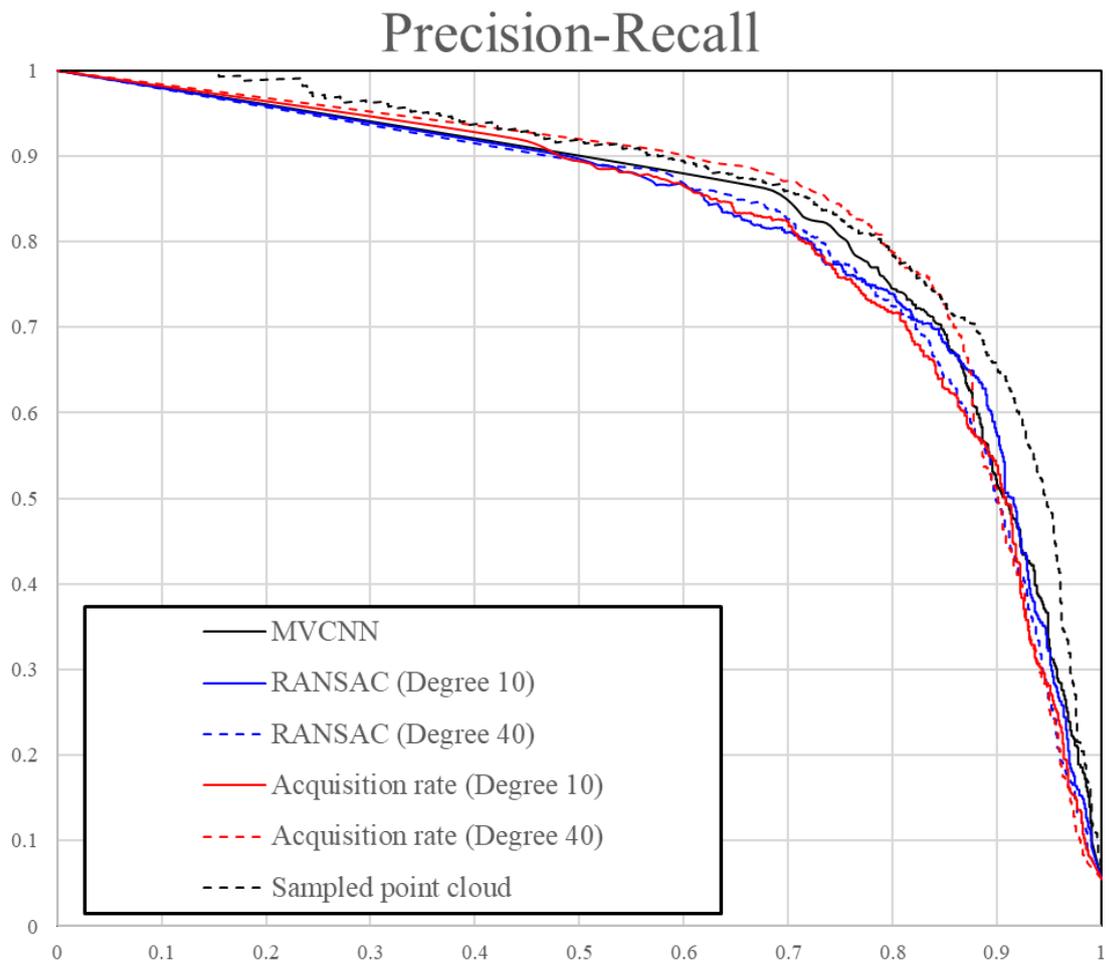

Figure 15. Precision-recall curve from six test cases

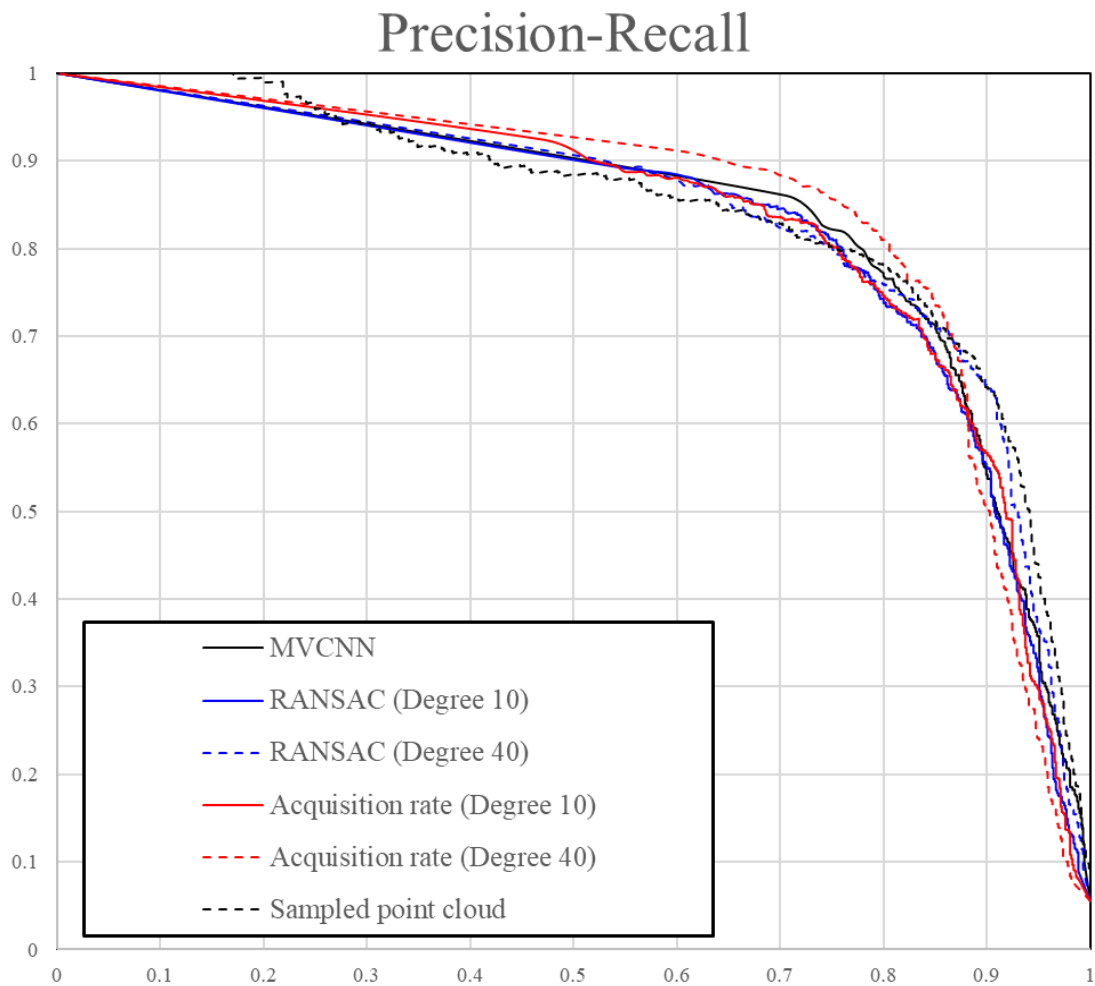

Figure 16. Precision-recall curve from six test cases when lower 50% types of fitting components that had a small amount of data were excluded

The MVCNN and PointNet networks showed similar performance but their performance differed depending on the characteristics of the fitting shapes. If sufficient training data can be acquired, the lower 50% types are expected to show high accuracy as well. Both networks showed excellent performance considering the loss of shapes information due to occlusion in the scanned point clouds of the process plant and the imbalance of point cloud density.

## 7. Conclusion

In this study, we experimented with the segmented point cloud retrieval performance for the fitting type components of a process plant based on deep learning technology. Experiments were performed using two networks: the MVCNN, which is a view-based approach, and PointNet, which is designed to use point cloud data as direct input. For the MVCNN, five methods for rendering images with point clouds were applied. For training data, the existing segmented point cloud repository [35] was used. When the amount of training data was sufficient, both the MVCNN and PointNet showed similar performance, but there were performance differences depending on the characteristics of the shapes. Some fitting types were misrecognized as other fittings with similar shapes. The MVCNN showed higher performance in the accuracy of the retrieval result, whereas PointNet showed higher performance in mAP calculated based on precision-recall curves.

Further research is required to investigate data augmentation to enhance performance for fitting types with a small amount of training data. Due to the characteristics of the environment, not all fitting components have similar quantities and the data collection cannot be performed consistently. Therefore, it is necessary to increase training data through data augmentation to achieve higher performance even with a small amount of data. Augmentation methods considering the characteristics of point cloud data should be researched, particularly for PointNet, which uses point cloud data as direct input.

**Acknowledgements**

This research was supported by the Industrial Core Technology Development Program (Project ID: 20000725) funded by the Korea government (MOTIE), and by the Basic Science Research Program (Project ID: NRF-2019R1F1A1053542) through the National Research Foundation of Korea (NRF) funded by the Korea government (MSIT).

**Conflict of Interest Statement**

On behalf of all authors, the corresponding author states that there is no conflict of interest.